\documentclass[10pt,twocolumn,letterpaper]{article}

\usepackage[pagenumbers]{cvpr} 

\usepackage{expl3}
\usepackage{atbegshi}
\usepackage{graphicx}
\usepackage{caption}
\usepackage{array}
\definecolor{cvprblue}{rgb}{0.21,0.49,0.74}
\usepackage[pagebackref,breaklinks,colorlinks,allcolors=cvprblue]{hyperref}

\def\paperID{4} 
\def\confName{CVPR}
\def\confYear{2025}

\title{Multi-Step Guided Diffusion for Image Restoration on Edge Devices: Toward Lightweight Perception in Embodied AI}

\author{Aditya Chakravarty\\
Independent Research\\
San Francisco, CA\\
{\tt\small chakravarty.aditya28@gmail.com}}

\AtBeginShipout{%
  \AtBeginShipoutFirst{%
    \vspace*{2mm}
    \par
    \centering
    {\normalsize Accepted in CVPR 2025 Embodied AI Workshop}
    \vspace*{2mm}
  }%
}

\begin{document}

\maketitle
\section{Introduction}

Diffusion models have emerged as powerful tools for solving inverse problems without task-specific retraining. Methods like Diffusion Posterior Sampling (DPS)~\cite{chung2023dps} and FreeDoM~\cite{yu2023freedom} guide the generative process using externally defined objectives, enabling flexible and modular inference. Related multi-call approaches~\cite{lugmayr2022repaint,song2023lgd,pan2023sag} extend these ideas through iterative guidance, but often require retraining, complex scheduling, or domain-specific tuning.

Manifold Preserving Guided Diffusion (MPGD)~\cite{he2024mpgd} was recently proposed as a training-free alternative that constrains guidance to the tangent space of a learned image manifold, improving stability and realism during restoration. However, MPGD and similar methods typically apply only a single gradient update per denoising step, leaving the potential of deeper, multi-step conditioning unexplored. Moreover, MPGD was developed and evaluated primarily on face-centric datasets, raising questions about its robustness on more diverse or out-of-distribution content.

In this work, we revisit MPGD through the lens of multi-step optimization: applying several gradient descent updates within each denoising timestep. Inspired by prior observations in RePaint~\cite{lugmayr2022repaint} and LGD~\cite{song2023lgd} that repeated updates can improve fidelity, we conduct an empirical study to probe the trade-offs between quality, diversity, and inference cost. We find that increasing the number of guidance steps significantly boosts both perceptual quality (LPIPS) and pixel-level accuracy (PSNR), while also enhancing robustness to degraded or out-of-distribution inputs. Notably, we show that MPGD—despite being trained on face datasets—can effectively restore generic, non-face natural images through multi-step conditioning.

Our experiments focus on two canonical inverse problems: $4\times$ super-resolution and Gaussian deblurring. We evaluate across a range of optimization depths using a Jetson Orin Nano, a compact edge GPU platform aligned with the practical constraints of embodied AI systems. Results on both ImageNet and UAV123 aerial imagery demonstrate that MPGD with multi-step optimization is a viable, lightweight solution for real-time visual restoration in robotics and mobile AI agents operating in unconstrained environments.

\vspace{-2mm}
\begin{figure*}
    \centering
    \scriptsize
    \begin{tabular}{>{\centering\arraybackslash}m{2.8cm} 
                    >{\centering\arraybackslash}m{2.8cm} 
                    >{\centering\arraybackslash}m{2.8cm} 
                    >{\centering\arraybackslash}m{2.8cm} 
                    >{\centering\arraybackslash}m{2.8cm}}
        \textbf{Input} & \textbf{1 Step} & \textbf{7 Steps} & \textbf{15 Steps} & \textbf{Ground Truth} \\
        \includegraphics[width=2.8cm]{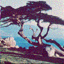} & 
        \includegraphics[width=2.8cm]{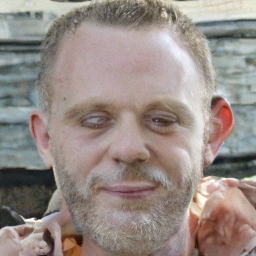} & 
        \includegraphics[width=2.8cm]{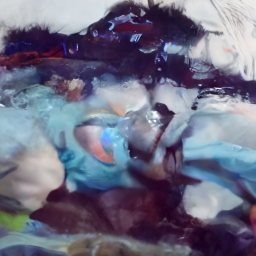} & 
        \includegraphics[width=2.8cm]{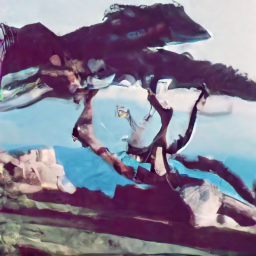} & 
        \includegraphics[width=2.8cm]{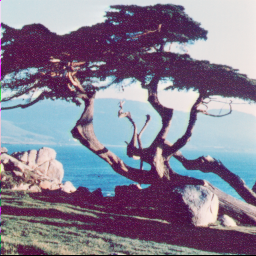} \\
        \includegraphics[width=2.8cm]{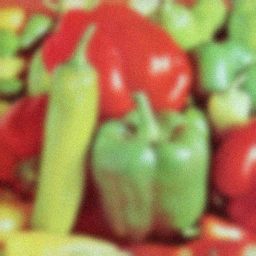} & 
        \includegraphics[width=2.8cm]{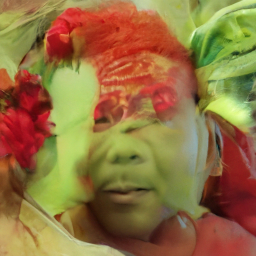} & 
        \includegraphics[width=2.8cm]{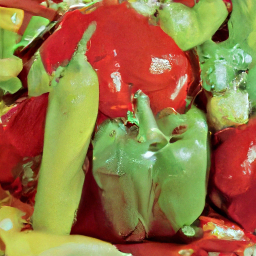} & 
        \includegraphics[width=2.8cm]{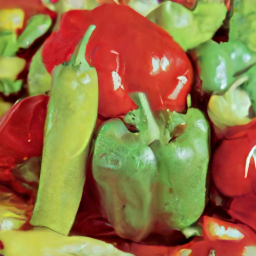} & 
        \includegraphics[width=2.8cm]{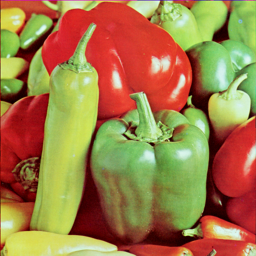} \\
    \end{tabular}
    \vspace{-1mm}
    \caption{\scriptsize Comparison of SR and Deblur results at 1, 7, and 15 steps.}
    \label{fig:qualitative}
    \vspace{-3mm}
\end{figure*}

\section{Method and Experimental Setup}

We consider $4\times$ super-resolution (bicubic downsampling) and Gaussian deblurring (kernel size 61, intensity 3.0), both with additive noise $\sigma=0.05$, following~\cite{he2024mpgd,wang2022ddnm}. We use the pixel-space MPGD implementation  with DDIM sampling~\cite{song2021ddim} and a pretrained FFHQ autoencoder~\cite{tov2021psp}.
 We generate outputs for 1000 ImageNet \cite{deng2009imagenet} test images, sweeping steps $\in \{1, 3, 7, 15, 20\}$ , timesteps $\in \{20, 50, 100\}$, and guidance scales $\in \{4, 7.5, 17.5\}$. For evaluation we consider LPIPS~\cite{zhang2018lpips}, SSIM~\cite{wang2004ssim}, inference time and PSNR. All experiments run on a single NVIDIA Jetson Orin Nano (8 GB VRAM).

\paragraph{Application Case: Aerial Inspection from UAV123}

To assess real-world viability in an embodied AI context, we evaluate MPGD on degraded aerial footage from the UAV123 dataset~\cite{uav123}. This dataset comprises UAV video sequences over buildings, roads, and industrial sites, which are representative of visual inspection scenarios. 

We extract 300 diverse frames spanning multiple scenes. These are processed using the same MPGD configuration as the main ImageNet benchmarks, without further training or adaptation. Outputs are compared across LPIPS, PSNR, SSIM, and inference time.

\section{Results and Discussion}

Across both tasks, we observe that performance improves with more optimization steps and moderate guidance scales, saturating around 15 steps. The reconstructions (Figure~\ref{fig:qualitative}
) evolve from generic face-like shapes at 1 step to well-structured outputs at 15 steps, even for out-of-distribution images. Inference latency ranges from 50–100ms per image on Jetson Orin Nano (Table~\ref{tab:mpgd_comparison}), validating MPGD’s applicability for real-time embedded perception.

\paragraph {Performance on UAV123 dataset} On UAV123 images, MPGD achieves strong perceptual and pixel-level performance, despite scene variability and noise. Table~\ref{tab:uav_case} shows MPGD (15 steps) surpasses NAFNet\cite{chen2022nafnet} and Uformer\cite{wang2022uformer} in LPIPS and PSNR while maintaining real-time throughput on Jetson Orin Nano. This suggests MPGD’s utility as a lightweight plug-in module for real-time image enhancement in aerial infrastructure inspection.

\begin{table}[h]
  \centering
  \scriptsize  
  \begin{tabular}{lcccc}
    \toprule
    Method & LPIPS $\downarrow$ & SSIM $\uparrow$ & PSNR $\uparrow$ & Time (ms) $\downarrow$ \\
    \midrule
    MPGD (15 steps) & 0.32 & 0.90 & 20.91 & 80 \\
    NAFNet~\cite{chen2022nafnet} & 0.36 & 0.86 & 20.13 & 35 \\
    Uformer~\cite{wang2022uformer} & 0.34 & 0.87 & 19.65 & 58 \\
    \bottomrule
  \end{tabular}
  \caption{Comparison of MPGD with baseline models (ImageNet).}
  \label{tab:mpgd_comparison}
\end{table}

\vspace{-3mm}
\begin{table}[h]
  \centering
  \scriptsize  
  \begin{tabular}{lcccc}
    \toprule
    Method & LPIPS $\downarrow$ & SSIM $\uparrow$ & PSNR $\uparrow$ & Time (ms) $\downarrow$ \\
    \midrule
    MPGD (15 steps) & 0.35 & 0.88 & 21.20 & 90 \\
    NAFNet & 0.38 & 0.84 & 20.10 & 42 \\
    Uformer & 0.37 & 0.85 & 19.90 & 65 \\
    \bottomrule
  \end{tabular}
  \caption{Performance on degraded UAV123 frames (300 samples).}
  \label{tab:uav_case}
\end{table}

Across both tasks, we observe that performance improves with more optimization steps and moderate guidance scales, saturating around 15 steps. Qualitatively, reconstructions evolve from generic face-like shapes at 1 step to well-structured outputs at 15 steps, even for out-of-distribution images. Inference latency ranges from 50–100ms per image, validating MPGD’s applicability for real-time embedded perception.

\section*{Conclusions and Ongoing/Future Work}

We presented a multi-step Manifold Preserving Guided Diffusion (MPGD) approach for training-free image restoration, targeting deployment on edge devices like the Jetson Orin Nano. Our experiments on both standard benchmarks and the UAV123 dataset demonstrate MPGD’s utility as a lightweight, retraining-free vision module for embodied AI agents, such as drones and mobile robots, which operate under stringent power and compute constraints. Crucially, we show that multi-step optimization allows a model trained on face-centric data to generalize surprisingly well to natural, non-face images—suggesting that task-driven optimization can compensate for mismatched training domains in practice.

This work is part of an ongoing effort to extend MPGD to a broader class of embodied perception challenges, including low-light navigation, infrastructure inspection, and visual localization under domain shift. Recent advances in diffusion-based decision-making~\cite{chi2023diffusion, liang2023generative} suggest that generative priors can support robust visual modules across diverse environments. Building on this, we plan to explore adaptive optimization depth during inference~\cite{song2023lgd}, as well as lightweight test-time adaptation strategies tailored to resource-constrained edge deployment~\cite{wang2023ttalite}. We are also investigating extensions of MPGD for multi-modal and non-linear inverse problems, including conditioning on spatial prompts or language cues. These directions aim to position MPGD as a deployable and adaptable backbone for real-time perception in embodied AI systems.


\begin{thebibliography}{18}
\providecommand{\natexlab}[1]{#1}
\providecommand{\url}[1]{\texttt{#1}}
\expandafter\ifx\csname urlstyle\endcsname\relax
  \providecommand{\doi}[1]{doi: #1}\else
  \providecommand{\doi}{doi: \begingroup \urlstyle{rm}\Url}\fi

\bibitem[Andreas Lugmayr et~al.(2022)Andreas Lugmayr, Martin Danelljan, Andres Romero, Fisher Yu, Radu Timofte, and Luc Van Gool]{lugmayr2022repaint}
Andreas Lugmayr, Martin Danelljan, Andres Romero, Fisher Yu, Radu Timofte, and Luc Van Gool.
\newblock {RePaint}: Inpainting using denoising diffusion probabilistic models.
\newblock In \emph{Proc.\ IEEE/CVF Conference on Computer Vision and Pattern Recognition (CVPR)}, 2022.

\bibitem[Chen et~al.(2022)Chen, Chu, and Zhang]{chen2022nafnet}
Liangyu Chen, Xiaojie Chu, and Xiangyu Zhang.
\newblock Simple baselines for image restoration.
\newblock In \emph{European Conference on Computer Vision (ECCV)}, pages 186--202, 2022.

\bibitem[Chi et~al.(2023)Chi, Huang, Yu, Ma, Handa, and Song]{chi2023diffusion}
Linxi Chi, Zichen Huang, Tao Yu, Ziyu Ma, Ankur Handa, and Shuran Song.
\newblock Diffusion policy: Visuomotor policy learning via action diffusion.
\newblock \emph{arXiv preprint arXiv:2303.04137}, 2023.

\bibitem[Chung et~al.(2023)Chung, Kim, McCann, L.Klasky, and Ye]{chung2023dps}
Hyungjin Chung, Jeongsol Kim, Michael~T. McCann, Marc L.Klasky, and JongChul Ye.
\newblock Diffusion posterior sampling for general noisy inverse problems.
\newblock In \emph{Proc.\ International Conference on Learning Representations (ICLR)}, 2023.

\bibitem[Deng et~al.(2009)Deng, Dong, Socher, Li, Li, and Fei-Fei]{deng2009imagenet}
Jia Deng, Wei Dong, Richard Socher, Li-Jia Li, Kai Li, and Li Fei-Fei.
\newblock Imagenet: A large-scale hierarchical image database.
\newblock \emph{Proceedings of the IEEE Conference on Computer Vision and Pattern Recognition (CVPR)}, pages 248--255, 2009.

\bibitem[He et~al.(2024)He, Murata, Lai, Takida, Uesaka, Kim, Liao, Mitsufuji, Kolter, Salakhutdinov, and Ermon]{he2024mpgd}
Yutong He, Naoki Murata, Chieh-Hsin Lai, Yuhta Takida, Toshimitsu Uesaka, Dongjun Kim, Wei-Hsiang Liao, Yuki Mitsufuji, J. Zico Kolter, Ruslan Salakhutdinov, and Stefano Ermon.
\newblock Manifold preserving guided diffusion.
\newblock In \emph{Proc.\ International Conference on Learning Representations (ICLR)}, 2024.
\newblock poster.

\bibitem[Jiachun Pan et~al.(2023)Jiachun Pan, Hanshu Yan, Jun Hao Liew, Jiashi Feng, and Vincent Y.\ F. Tan]{pan2023sag}
Jiachun Pan, Hanshu Yan, Jun Hao Liew, Jiashi Feng, and Vincent Y.\ F. Tan.
\newblock Towards accurate guided diffusion sampling through symplectic adjoint method.
\newblock \emph{arXiv preprint arXiv:2312.12030}, 2023.

\bibitem[Liang et~al.(2023)Liang, Yuan, Wu, et~al.]{liang2023generative}
Jacky Liang, Zirui Yuan, Qiyang Wu, et~al.
\newblock Generative pretraining for decision making and control.
\newblock \emph{arXiv preprint arXiv:2210.03029}, 2023.

\bibitem[Mueller et~al.(2016)Mueller, Smith, and Ghanem]{uav123}
Matthias Mueller, Neil Smith, and Bernard Ghanem.
\newblock A benchmark and simulator for uav tracking.
\newblock In \emph{ECCV}, 2016.

\bibitem[Song et~al.(2021)Song, Meng, and Ermon]{song2021ddim}
Jiaming Song, Chenlin Meng, and Stefano Ermon.
\newblock Denoising diffusion implicit models.
\newblock In \emph{International Conference on Learning Representations (ICLR)}, 2021.

\bibitem[Song et~al.(2023)Song, Meng, and Ermon]{song2023lgd}
Yang Song, Chenlin Meng, and Stefano Ermon.
\newblock Loss-guided diffusion: Learning to denoise images conditioned on a loss function.
\newblock In \emph{International Conference on Machine Learning (ICML)}, 2023.

\bibitem[Tov et~al.(2021)Tov, Alaluf, Nitzan, Patashnik, and Cohen-Or]{tov2021psp}
Omer Tov, Yuval Alaluf, Yotam Nitzan, Or Patashnik, and Daniel Cohen-Or.
\newblock Designing an encoder for stylegan image manipulation.
\newblock In \emph{Proceedings of the IEEE/CVF Conference on Computer Vision and Pattern Recognition (CVPR)}, pages 17968--17977, 2021.

\bibitem[Wang et~al.(2004)Wang, Bovik, Sheikh, and Simoncelli]{wang2004ssim}
Zhou Wang, Alan~C Bovik, Hamid~R Sheikh, and Eero~P Simoncelli.
\newblock Image quality assessment: From error visibility to structural similarity.
\newblock \emph{IEEE Transactions on Image Processing}, 13\penalty0 (4):\penalty0 600--612, 2004.

\bibitem[Wang et~al.(2021)Wang, Cun, Bao, Chen, Gu, Liu, and Dong]{wang2022uformer}
Zhendong Wang, Xiaodong Cun, Jianmin Bao, Guihua Chen, Jin Gu, Jianzhuang Liu, and Chen Dong.
\newblock Uformer: A general u-shaped transformer for image restoration.
\newblock In \emph{Proceedings of the IEEE/CVF International Conference on Computer Vision (ICCV)}, pages 17683--17693, 2021.

\bibitem[Wang et~al.(2023)Wang, Wang, Zhang, Xu, and Wang]{wang2023ttalite}
Zhen Wang, Qiang Wang, Kai Zhang, Mengyuan Xu, and Yiran Wang.
\newblock Tta-lite: Memory-efficient test-time adaptation for deep models on the edge.
\newblock In \emph{Proceedings of the IEEE/CVF Conference on Computer Vision and Pattern Recognition (CVPR)}, 2023.

\bibitem[Yinhuai Wang et~al.(2022)Yinhuai Wang, Jiwen Yu, and Jian Zhang]{wang2022ddnm}
Yinhuai Wang, Jiwen Yu, and Jian Zhang.
\newblock Zero-shot image restoration using denoising diffusion null-space model.
\newblock \emph{arXiv preprint arXiv:2212.00490}, 2022.

\bibitem[Yu et~al.(2023)Yu, Wang, Zhao, Ghanem, and Zhang]{yu2023freedom}
Jiwen Yu, Yinhuai Wang, Chen Zhao, Bernard Ghanem, and Jian Zhang.
\newblock {FreeDoM}: Training-free energy-guided conditional diffusion model.
\newblock In \emph{Proc.\ IEEE/CVF International Conference on Computer Vision (ICCV)}, 2023.

\bibitem[Zhang et~al.(2018)Zhang, Isola, Efros, Shechtman, and Wang]{zhang2018lpips}
Richard Zhang, Phillip Isola, Alexei~A. Efros, Eli Shechtman, and Oliver Wang.
\newblock The unreasonable effectiveness of deep features as a perceptual metric.
\newblock In \emph{Proc.\ IEEE Conference on Computer Vision and Pattern Recognition (CVPR)}, pages 586--595, 2018.

\end{thebibliography}

{
    \small
    \bibliographystyle{ieeenat_fullname}

}


\end{document}